\documentclass{article}
\usepackage[utf8]{inputenc}
\usepackage{multirow}
\usepackage{booktabs}
\usepackage{graphicx}
\usepackage{subfig}
\usepackage{tablefootnote}
\usepackage{color}
\usepackage{mdwlist}
\usepackage[dvipsnames]{xcolor}
\usepackage{soul}
\usepackage{hyperref}
\usepackage{authblk}

\title{Keeping the Bad Guys Out: \\Protecting and Vaccinating Deep Learning with JPEG Compression}

\author[1]{Nilaksh Das}
\author[1]{Madhuri Shanbhogue}
\author[1]{Shang-Tse Chen}
\author[1]{Fred Hohman}
\author[2]{Li Chen}
\author[2]{Michael E. Kounavis}
\author[1]{Duen Horng Chau}
\affil[1]{Georgia Institute of Technology}
\affil[2]{Intel Corporation}

\date{}

\begin{document}

\maketitle

\section*{Abstract}
Deep neural networks (DNNs) have achieved great success in solving a variety of machine learning (ML) problems, especially in the domain of image recognition.
However, recent research showed that DNNs can be highly vulnerable to adversarially generated instances, which look seemingly normal to human observers, but completely confuse DNNs.
These adversarial samples are crafted by adding small perturbations to normal, benign images.
Such perturbations, while imperceptible to the human eye, are picked up by DNNs and cause them to misclassify the manipulated instances with high confidence.
In this work, we explore and demonstrate
how systematic JPEG compression can work as an effective pre-processing step in the classification pipeline to  counter adversarial attacks and dramatically reduce their effects (e.g., Fast Gradient Sign Method, DeepFool).
An important component of JPEG compression is its ability to remove high frequency signal components, inside square blocks of an image.
Such an operation is equivalent to selective blurring of the image, helping remove additive perturbations.
Further, we propose an ensemble-based technique that can be constructed quickly from a given well-performing DNN, and empirically show how such an ensemble that leverages JPEG compression can protect a model from multiple types of adversarial attacks, without requiring knowledge about the model.

\section{Introduction}

Over the past few years, deep neural networks have achieved huge success in many important applications.
Computer vision, in particular, enjoys some of the biggest improvement over traditional methods~\cite{krizhevsky2012imagenet}.
As the DNN models become more powerful, people tend to do less data pre-processing or manual feature engineering, and prefer  so-called end-to-end learning.
For example, instead of manual feature normalization or standardization, one can add batch normalization layers and learn the best way to do it from the data distribution~\cite{IoffeS15}.
Image denoising can also be performed by stacking a DNN on top of an auto-encoder~\cite{gu2014towards}.
\\ \\
However, recent research has shown serious potential vulnerability in DNN models~\cite{Szegedy14}, which demonstrates that adding some small and human-imperceptible perturbations on an image can mislead the prediction of a DNN model to some arbitrary class. These perturbations can be computed by using the gradient information of a DNN model, which guides the direction in the input space that will most drastically change the model outputs~\cite{goodfellow2014explaining,Papernot16limitation}.
To make the vulnerability even more troubling, it is possible to compute a single ``universal'' perturbation that can be applied to any images and mislead the classification results of the model~\cite{Moosavi17}. Also, one can perform black-box attacks without knowing the exact DNN model being used~\cite{Papernot17blackbox}.
\\ \\
Many defense methods have been proposed to counteract the adversarial attacks.
A common way is to design new network architectures or optimization techniques~\cite{gu2014towards,goodfellow2014explaining}.
However, finding a good network architecture and hyperparameters for a particular dataset can be hard, and the resulting model may only be resistant to certain kind of attacks.

\subsection{Our Contributions}

In this work, we propose to use JPEG compression as a simple and effective pre-processing step to remove adversarial noise.
Our intuition is that as adversarial noises are often indiscernible by the human eye, JPEG compression --- designed to selectively discard information unnoticeable to humans --- have strong potential in combating such manipulations.
\\ \\
Our approach has multiple desired advantages. First, JPEG is a widely-used encoding technique and many images are already stored in the JPEG format. Most operating systems also have built-in support to encode and decode JPEG images, so even non-expert users can easily apply this pre-processing step. Second, this approach does not require knowledge about the model nor the attack, and can be applied to a wide range of image datasets.
\\ \\
This work presents the following contributions:

\begin{itemize}

\item A pre-processing step to neural network image classifiers that uses JPEG compression to remove adversarial noise from a given dataset.

\item Empirical tests on two datasets, CIFAR-10 and GTSRB, that systematically studies how varying JPEG compression qualities affects prediction accuracy.

\item Results showing the effect of including various amount of JPEG compressed images in the training process. We find that this significantly boosts accuracies on adversarial images and does not hurt the performance on benign images.
\end{itemize}

\begin{figure*}[t]
\includegraphics[width=0.9\textwidth]{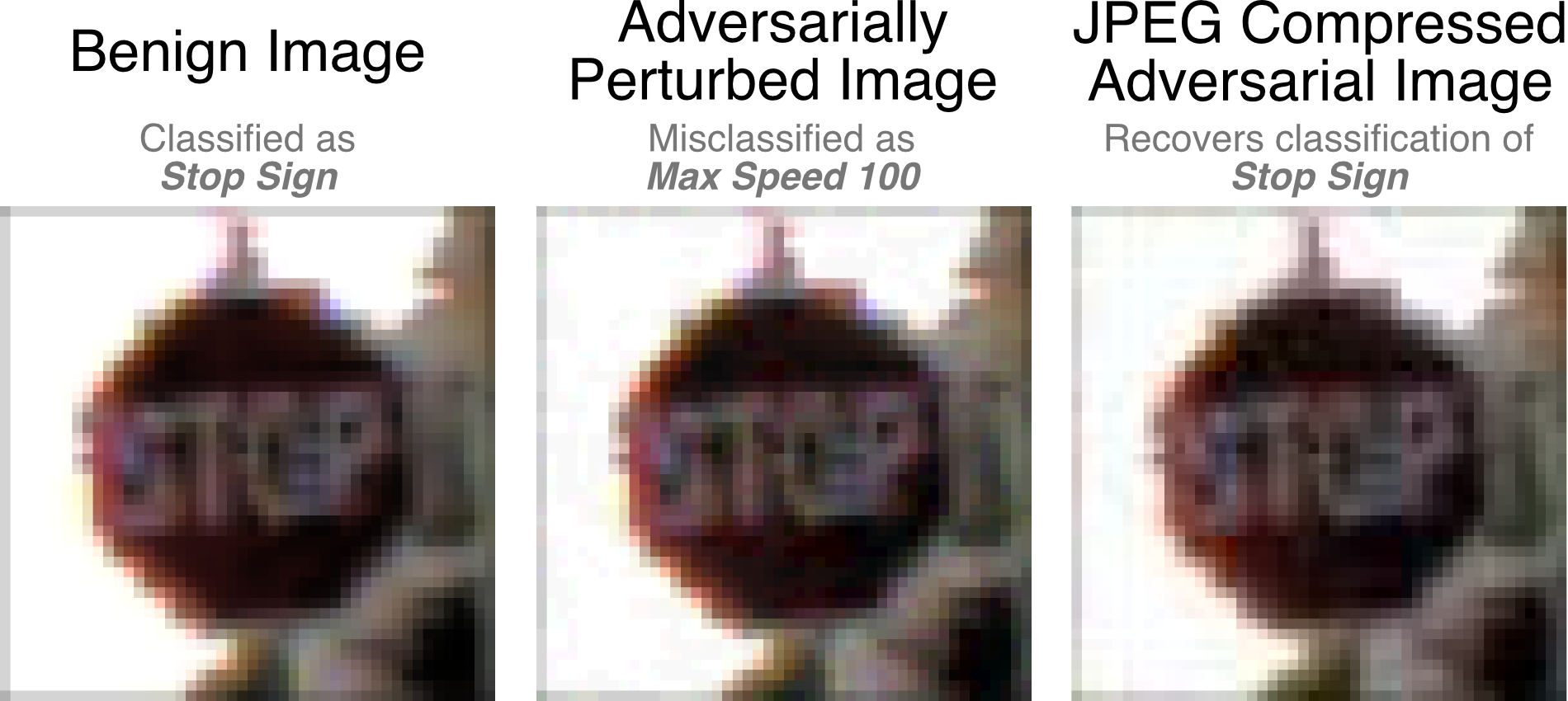}
\centering
\caption{A comparison of the classification results of an exemplar image from the German Traffic Sign Recognition Benchmark (GTSRB) dataset. A benign image (left) is originally classified as a \textit{stop sign}, but after the addition of an adversarial perturbation to the image (middle) the resulting image is classified as a \textit{max speed 100} sign. Using JPEG compression on the adversarial image (right), we recover the original classification of \textit{stop sign}.}
\label{fig:crown-jewel}
\end{figure*}

\section{Background}
In this section, we discuss existing adversarial attack algorithms and defense mechanisms.
We then give a brief overview of JPEG compression, which plays a crucial role in our defense approach.

\subsection{Adversarial Attacks}
Consider the scenario where a trained machine learning classifier $C$ is deployed.
An attacker, assumed to have full knowledge of the classifier $C$,
tries to compute a small distortion $\delta x$ for some test example $x$ such that
the perturbed example $x' = x + \delta x$ is misclassified by the model, i.e., $C(x') \ne C(x)$.
Prior work has shown that even if the machine learning model is unknown, one can train a substitute model and use it to compute the perturbation.
This approach is very effective in practice when both the target model and the substitute model are deep neural networks, due to the property of transferability~\cite{Szegedy14, Papernot17blackbox}.
\\ \\
The seminal work by Szegedy et al.~\cite{Szegedy14} proposed the first effective adversarial attack on DNN image classifiers by solving a Box-constrained L-BFGS optimization problem and showed that the computed perturbations to the images were indistinguishable to the human eye: a rather troublesome property for people trying to identify adversarial images. This discovery has gained tremendous interest, and many new attack algorithms have been invented~\cite{goodfellow2014explaining, Moosavi16, Moosavi17, Papernot16limitation} and applied to other domains such as malware detection~\cite{grosse2016malware, hu2017generating}, sentiment analysis~\cite{PapernotMSH16}, and reinforcement learning~\cite{lin2017tactics, huang2017adversarial}.
\\ \\
In this paper, we explore the \emph{Fast Gradient Sign Method} (\emph{FGSM})~\cite{goodfellow2014explaining} and the \emph{DeepFool} (\emph{DF})~\cite{Moosavi16} attack, each of which construct instance-specific perturbations to confuse a given model.
We choose the \emph{FGSM} attack for our evaluation since it is the most efficient approach in terms of computation time and the \emph{DF} attack because it computes minimal (i.e., highly unnoticeable) perturbations.
\\ \\
Since we plan to evaluate these attacks in a practical setting, we only consider the parameters of these attacks that produce adversarial instances having low pathology, i.e., the images are not obviously perceivable as having been manipulated, as seen in the middle image in Figure \ref{fig:crown-jewel}. Below, we briefly review the two attacks.
\\ \\
\textbf{Fast Gradient Sign Method}~\cite{goodfellow2014explaining}. \emph{FGSM} is a fast algorithm which computes perturbations subject to an $L_{\infty}$ constraint. The perturbation is computed by linearizing the loss function $J$,
$$
x' = x + \epsilon \cdot sign(\nabla J_x (\theta, x, y))
$$
where $\theta$ is the set of parameters of the model and $y$ is the true label of the instance. The parameter $\epsilon$ controls the magnitude of the perturbation.
Intuitively, this method uses the gradient of the loss function to determine in which direction each pixel’s intensity should be changed to minimize the loss function, and updates all pixels accordingly by a specific magnitude.
It is important to note here that \emph{FGSM} was designed to be a computational fast attack rather than an optimal attack. Therefore, it is not meant to produce minimal adversarial perturbations.
\\ \\
\textbf{DeepFool}~\cite{Moosavi16}. \emph{DF} constructs an adversarial instance under an $L_2$ constraint by assuming the decision boundary to be hyperplanar. The authors leverage this simplification to compute a minimal adversarial perturbation that results in a sample that is close to the original instance but orthogonally cuts across the nearest decision boundary. In this respect, \emph{DF} is an untargeted attack. Since the underlying assumption that the decision boundary is completely linear in higher dimensions is an oversimplification of the actual case, \emph{DF} keeps reiterating until a true adversarial instance is found.
The resulting perturbations are harder for humans to detect compared to perturbations applied by other techniques.

\subsection{Defense Mechanisms}
Although making a DNN model completely immune to adversarial attacks is still an open problem, there have been various attempts to mitigate the threat.
We summarize the approaches with four categories.
\begin{enumerate}
\item \textbf{Detecting adversarial examples before performing classification.}

Metzen et al.~\cite{metzen2017detecting} propose to distinguish genuine examples from the adversarially perturbed ones by augmenting deep neural networks with a small ``detector'' subnetwork. Feinman et al.~\cite{feinman2017detecting} use density estimates to detect examples that lie far from the natural data manifold, and use Bayesian uncertainty estimates to detect when examples lie in the low-confidence regions.

\item \textbf{Modifying network architecture.}
Deep Contractive Network~\cite{gu2014towards} is a generalization of the contractive autoencoder, which imposes a layer-wise contractive penalty in a feed-forward neural network. This approximately minimizes the network outputs variance with respect to perturbations in the inputs. Dense Associative Memory model~\cite{krotov2017dense}
tries to enforce higher order interactions between neurons by changing rectified linear unit (ReLU) to rectified polynomials. The idea is inspired by the hypothesis that adversarial examples are caused by high-dimensional linearity of DNN models.

\item \textbf{Modifying the training process.}
The most common and straightforward approach is to directly use adversarial examples to augment the training set. However, this is computationally expensive.
Goodfellow et al.~\cite{goodfellow2014explaining} simulate this process in a more efficient way by using a modified loss function that takes a perturbed example into account. Papernot et al.~\cite{papernot2016distillation} use the distillation method that uses the soft outputs of the first model as labels to train a second model.

\item \textbf{Pre-processing input examples to remove adversarial perturbation.}
A major advantage of this approach is that it can be used with any machine learning model, therefore it can be used alongside any other method described above. Bhagoji et al.~\cite{bhagoji2017dimensionality} apply principal component analysis on images to reduce dimension and discard noise.
Luo et al.~\cite{luo2015foveation} propose to use a foveation-based mechanism that applies a DNN model on a certain region of an image and discards information from other regions.
\\ \\
Our work belongs to this category. Prior works most relevant to our proposed method are \cite{dziugaite2016study} and \cite{kurakin2016adversarial}, both of which include JPEG compression as their defense mechanism.
However, previous work did not focus on how JPEG compression may be systematically leveraged as a defense mechanism.
For example, \cite{dziugaite2016study} only studied JPEG compression with quality 75 and had not evaluated how varying the amount of compression would affect performance.
In this work, we conduct extensive study to understand the compression approach's capability.
\\ \\
\cite{kurakin2016adversarial} tried pre-processing techniques besides JPEG compression, such as printing out adversarial images and taking pictures of them using a cell phone, changing contrast, and adding Gaussian noise.
However, they only processed the images during the testing phase.
In contrast, we also consider training our models with JPEG compressed images.
We further show that constructing an ensemble of the models obtained by training on images of different levels of compression quality can  significantly boost the success rates in recovering the correct answers in adversarial images.

\end{enumerate}

\subsection{JPEG compression}

JPEG is a standard and widely-used image encoding and compression technique consists of the following steps:
\begin{enumerate}
    \item converting the given image from \emph{RGB} to \emph{Y$C_bC_r$} color space: this is done because the human visual system relies more on spatial content and acuity than it does on color for interpretation. Converting the color space isolates these components which are of more import.

    \item performing spatial subsampling of the chrominance channels in the \emph{Y$C_bC_r$} space: the human eye is much more sensitive to changes in luminance, and downsampling the chrominance information does not affect the human perception of the image very much.

    \item transforming a blocked representation of the \emph{Y$C_bC_r$} spatial image data to a frequency domain representation using Discrete Cosine Transform (DCT): this step allows the JPEG algorithm to further compress the image data as outlined in the next steps by computing DCT coefficients.

    \item performing quantization of the blocked frequency domain data according to a user defined quality factor: this is where the JPEG algorithm achieves majority of the compression, at the expense of image quality. This step suppresses higher frequencies more since these coefficients contribute less to the human perception of the image.

\end{enumerate}

\section{Experimental setup}

Experiments in this paper were conducted with convolutional neural networks on two image datasets: the \textit{CIFAR-10} dataset \cite{krizhevsky2009cifar10}, and the \textit{German Traffic Sign Recognition Benchmark} (GTSRB) dataset \cite{Stallkamp-IJCNN-2011}.
\\ \\
The CIFAR-10 dataset consists of 50,000 training examples and a test set of 10,000 examples with 10 classes. Each image in the dataset is of size $32 \times 32$ pixels. The GTSRB dataset has 43 classes with 39,209 training examples and 12,630 testing examples. The image sizes in this dataset vary between $15 \times 15$ to $250 \times 250$ pixels. For our analysis, we rescale each image to $48 \times 48$ pixels.
\\ \\
For CIFAR-10, we use a convolutional neural network with 2 Conv-Conv-Pooling blocks, having Conv layer filter depths of 32 and 64 respectively.
The Conv and Pooling filter size used are $3 \times 3$ with a Pooling stride of 2, 2.
This is followed by a fully connected layer of 512 units that feeds into a softmax output layer of 10 classes.
The same architecture is extended for the GTSRB dataset with an additional Conv-Conv-Pooling block of filter depth 128 and a softmax output layer of 43 classes. The Pooling filter size is made $2 \times 2$.
\\ \\
Both model was trained for 400 epochs using categorical cross entropy loss with dropout regularization. We used the Adam optimizer to find the best weights. The final models obtained had testing accuracy of 82.88\% and 97.83\% on CIFAR-10 and GTSRB respectively.
\\ \\
To measure the effectiveness of an adversarial attack, we use a metric that we call the ``misclassification success" rate. It is defined as the proportion of instances which were correctly classified by the trained models and whose labels were successfully flipped by the attack.

\section{JPEG Compression as Defense}

A core principle behind JPEG compression is based on the human psychovisual system, which aims to suppress high frequency information like sharp transitions in intensity and color hue using Discrete Cosine Transform.
As adversarial attacks often introduce perturbations that are not compatible with human psychovisual awareness (hence these attacks are sometimes imperceptible to humans), and we believe JPEG compression has the potential to remove these artifacts.
Thus, we propose to use JPEG compression as a pre-processing step before running an instance through the classification model.
We demonstrate how using JPEG compression reduces the mistakes a model makes on datasets that have been adversarially manipulated.

\begin{figure*}[t]
\includegraphics[width=1.0\textwidth]{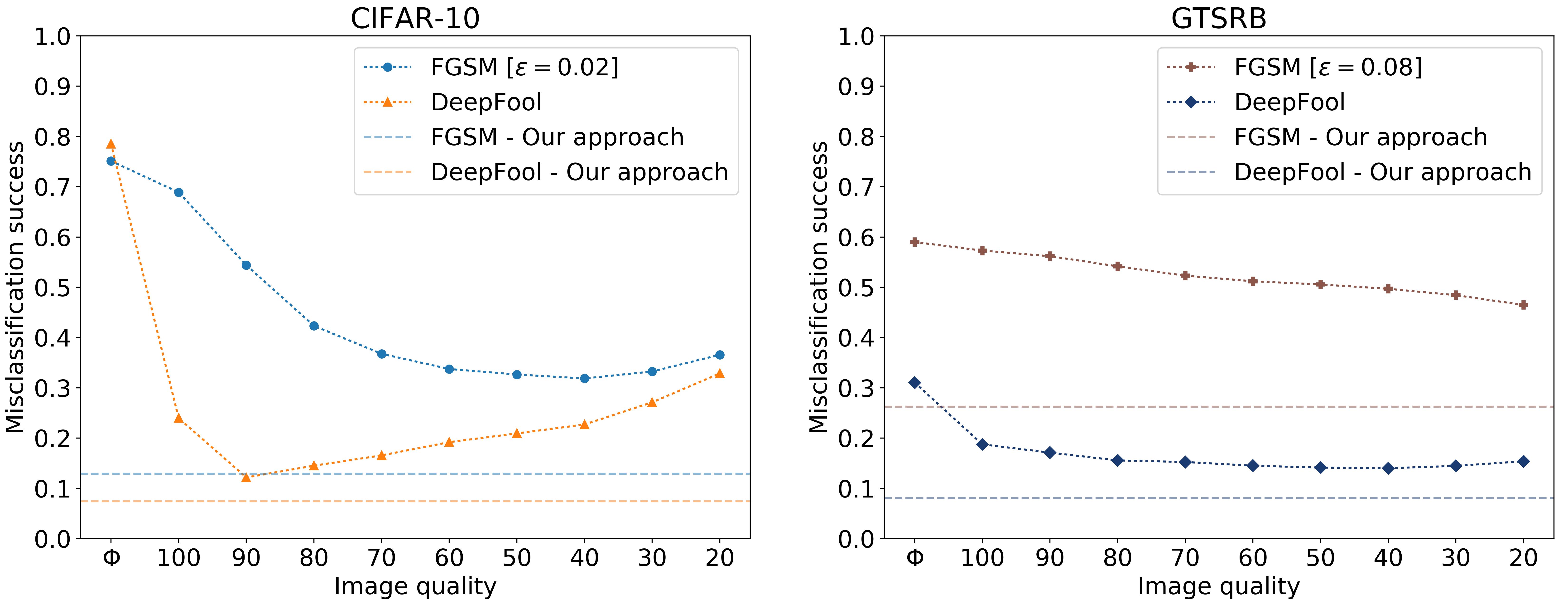}
\centering
\caption{Applying JPEG compression (dashed lines with symbols) can counter \emph{FGSM} and \emph{DeepFool} attacks on the CIFAR-10 and GTSRB datasets, e.g., slightly compressing CIFAR-10 images  dramatically lowers \emph{DeepFool}'s attack success rate, indicated by the steep orange line (left plot).
$\Phi$ means no compression has been applied.
Attacks can be further suppressed by ``vaccinating'' a DNN model by training it with compressed images, and using an ensemble of such models -- our approach, discussed in Section \ref{sec:our-approach},
rectifies a great majority of misclassification (indicated by the horizontal dashed lines).
}
\label{fig:misclass-4_1}
\end{figure*}

\subsection{Effect of JPEG Compression on Classification}
Benign, everyday images lie in a very narrow manifold.
An image with completely random pixel colors is highly unlikely to be perceived as natural  by  human beings.
However, the objective basis of classification models, like DNNs, often are not aligned with such considerations.
DNNs may be viewed as constructing decision boundaries that linearly separates the data in high dimensional spaces.
In doing so, these models assume that the subspaces of natural images exist beyond the actual manifold.
Adversarial attacks take advantage of this by perturbing images just enough so that they cross over the decision boundary of the model.
However, this crossover does not guarantee that the perturbed images would lie in the original narrow manifold.
Indeed, perturbed images could lie in  artificially expanded subspaces where natural images would not be found.
\\ \\
Since JPEG compression takes the human psychovisual system into account, we pursue the hypothesis that the manifold in which JPEG images occur would have some semblance with the manifold of naturally occurring images, and that using JPEG compression as a pre-processing step during classification would re-project any adversarially perturbed instances back onto this manifold.
\\ \\
To test our hypothesis, we applied JPEG compression to images from the CIFAR-10 and GTSRB datasets, adversarially perturbed by \emph{FGSM} and \emph{DF}, and varied the quality parameter of the JPEG algorithm. Figure \ref{fig:misclass-4_1} shows the experiment results.
\\ \\
Overall, we observe that applying JPEG compression (dashed lines with symbols) can counter \emph{FGSM} and \emph{DeepFool} attacks on the CIFAR-10 and GTSRB datasets.
$\Phi$ means no compression has been applied.
\\ \\
Increasing compression (decreasing image quality) generally leads to better removal of the adversarial effect at first, but the benefit reaches an inflection point where the success rate starts increasing again.
Besides the adversarial perturbation, this inflection may also be attributed to the artifacts introduced by JPEG compression itself at lower image qualities, which confuses the model.
\\ \\
With CIFAR-10, we observe that slightly compressing its images dramatically lowers \emph{DeepFool}'s attack success rate, indicated by the steep orange line (left plot).
The steepest drops take place on applying JPEG compression of image quality 100 on uncompressed images, introducing extremely little compression in the frequency domain.
Since the JPEG algorithm also performs downsampling of the chrominance channel irrespective of the image quality, a hypothesis that supports this observation may be that \emph{DF} attacks the chrominance channel much more than the luminance channel in $YC_bC_r$ color space.
Since \emph{DF} introduces a minimal perturbation, it is easily removed with JPEG compression.

\subsection{Vaccinating Models by Training with JPEG Compressed Images}

\begin{figure*}[!h]
\includegraphics[width=0.5\textwidth]{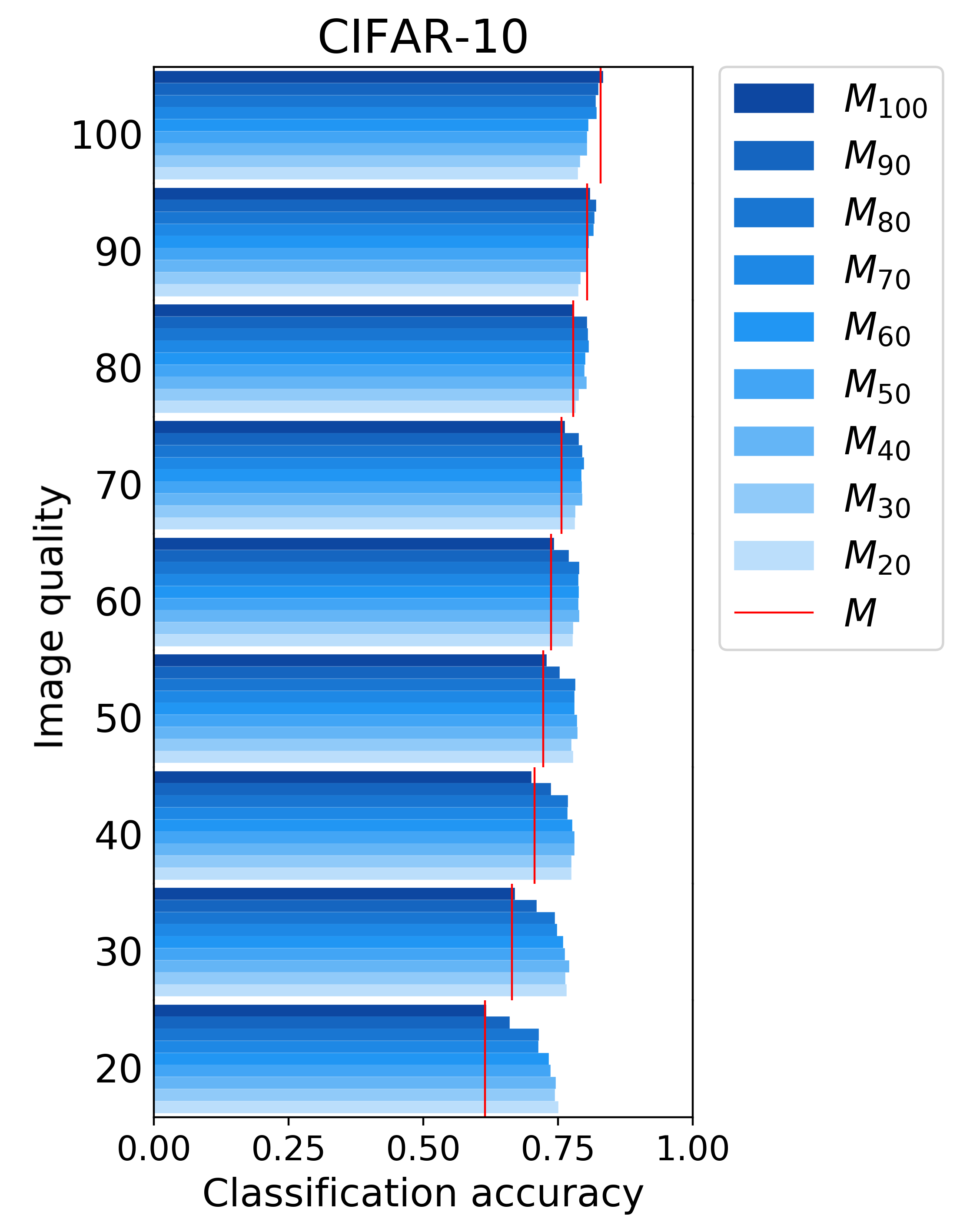}
\centering
\caption{Classification accuracies of each vaccinated model on the CIFAR-10 test set that has been compressed to a particular image quality. Each cluster of bars represents the model performances when tested with images having the corresponding image quality as indicated on the vertical axis.
Within each cluster, each bar represents a vaccinated model.
Vertical red lines denote the accuracy of the original, non-vaccinated model for that image quality.}
\label{fig:retracc-4_2}
\end{figure*}

\noindent Testing adversarial images with JPEG compression suggests that the algorithm seems to be able to remove perturbations by re-projecting the images to the manifold of JPEG images.
Since our initial model was trained on the original benign image dataset (without any adversarial manipulation), testing with compressed images that have lower image quality unsurprisingly lead to higher misclassification rates, likely due to artifacts introduced by the compression algorithm itself.
This can also be explained by the notion that the manifold of JPEG compressed images of a particular image quality may be similar to that of another quality, but not completely aligned.
We now propose that with training the model over this manifold corresponding to a particular image quality, the model can potentially learn to classify images even in the presence of JPEG artifacts.
From the perspective of adversarial images, applying JPEG compression would remove the perturbations and re-training with compressed images could help ensure that the model is not confused by the JPEG artifacts.
We call this approach of re-training the model with JPEG compressed images as ``\textbf{vaccinating}" the model against adversarial attacks.
\\ \\

\begin{figure*}[t]
\includegraphics[width=1.0\textwidth]{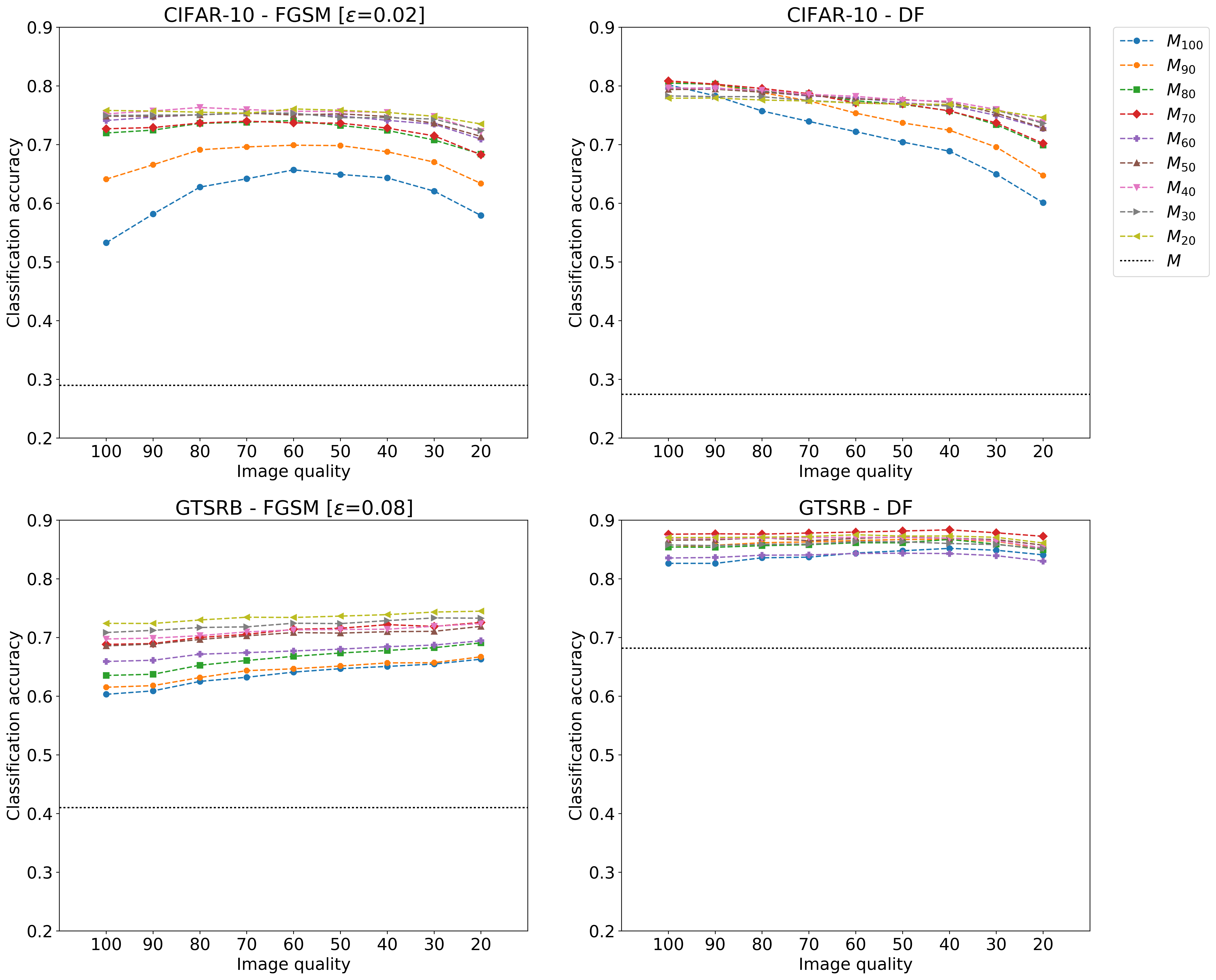}
\centering
\caption{Performance of the vaccinated models on adversarially constructed test sets. Each line with a symbol represents a vaccinated model and the black horizontal dotted line represents the accuracy of the original model under attack.
These results demonstrate that re-training with JPEG compressed images can help recover from an adversarial attack.}
\label{fig:attackacc-4_2}
\end{figure*}

\noindent We re-trained the model with images of JPEG qualities 100 through 20 (increasing compression) with a step size of 10, and hence obtained 9 models (besides the original model). We refer to each of these re-trained models as $M_x$, where $x$ corresponds to the image quality the model was re-trained with. The original model is referred to as $M$.
\\ \\
While re-training, the weights of $M_x$ were initialized with the weights of $M_{x+10}$ for faster convergence.
For example, the weights of $M_{80}$ were initialized with weights of $M_{90}$, and the weights of $M_{100}$ were initialized with weights of $M$, and so on.
The intuition for our approach was derived from the proposition that the manifold of images corresponding to successive levels of compression would exist co-locally, and the decision boundaries learned by the model would not have to displace significantly to account for the new manifold. This means that given any model, our approach can quickly generate new vaccinated models.
\\ \\
Figure \ref{fig:retracc-4_2} shows clear benefits of our  vaccination idea --- vaccinated models generally perform better than the original model on the CIFAR-10 test set, especially at lower image qualities.
For example, $M_{20}$ performs the best for images with quality 20 and worst for images with quality 100. Correspondingly, $M_{100}$ performs the best for images with quality 100 and worst for images with quality 20.
The performance of $M_{100}$ closely follows the performance of $M$ across each image quality.
All these observations are consistent with our fundamental intuition of JPEG manifolds coexisting in the same hyperlocality.
\\ \\
Figure \ref{fig:attackacc-4_2} visualizes the performance of the vaccinated models on adversarially perturbed datasets by varying the image quality it is tested on.
Again, general trends show that increasing JPEG compression removes adversarial perturbations. We see that the effect of the adversarial attacks on $M$ does get transferred to the vaccinated models as well, but as the compression is increased on the images that the model is trained with, the transferability of the attack subsides. An interesting thing to note here is that with CIFAR-10, the accuracy decreases for lower image qualities. This means that the artifacts introduced by JPEG may be taking over the adversarial attack to bring down the accuracy, which may be attributed to the small image size of the CIFAR-10 dataset.
We do not see such a significant decrease in accuracy at lower image qualities with the GTSRB dataset, which contains larger images.

\section{Fortified Defense: an Ensemble of Models} \label{sec:our-approach}

\begin{figure*}[!h]
\includegraphics[width=1.0\textwidth]{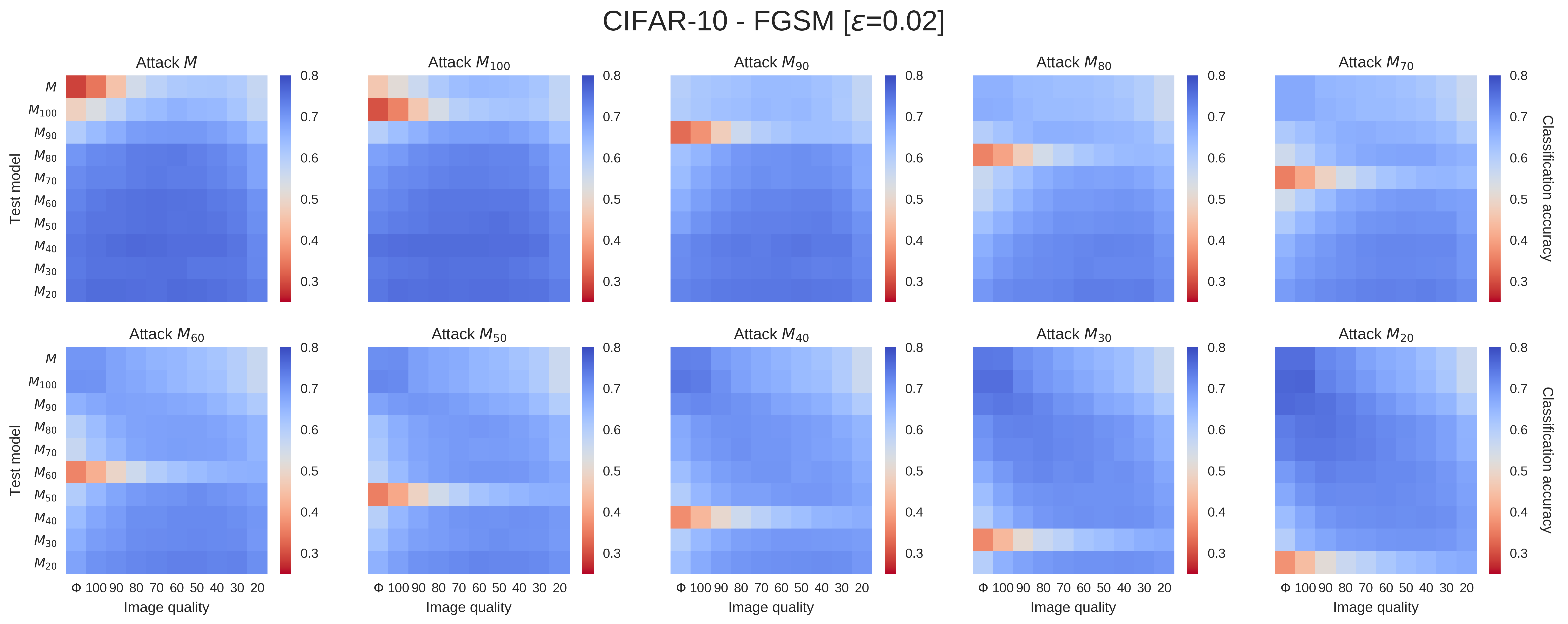}
\centering
\caption{Accuracies of all models under consideration when each model is individually attacked. The attack does get transferred to other models but is mitigated with increasing JPEG compression.}
\label{fig:transferability-5}
\end{figure*}

\begin{table}[b]
\centering
\caption{Performance of our approach on the respective test sets as compared to the scenario when original, non-vaccinated model is under attack.}
\label{tbl:accs-5}
\begin{tabular}{l l rr}
\toprule
                          &                         & \multicolumn{1}{c}{Original scenario} & \multicolumn{1}{c}{With our ensemble} \\ \midrule
\multirow{3}{*}{CIFAR-10} & Benign images           & 82.88\%                               & 83.19\%                               \\
                          & FGSM [$\epsilon$=0.02]  & 28.97\%                               & 79.57\%                               \\
                          & DeepFool                & 27.44\%                               & 82.71\%                               \\ \midrule
\multirow{3}{*}{GTSRB}    & Benign images           & 97.83\%                               & 98.59\%                               \\
                          & FGSM [$\epsilon$=0.08]  & 41.00\%                                & 73.37\%                               \\
                          & DeepFool                & 68.19\%                               & 91.70\%                               \\ \bottomrule
\end{tabular}
\end{table}

If adversaries are able to gain access to the original model, they may also be able to recreate the vaccinated models and attack them individually.
To protect the classification pipeline against such an attack, we propose to use an ensemble of vaccinated models that vote on images with varying image qualities.
Hence, in our ensemble, the models $M_{100}$ through $M_{20}$ vote on a given image compressed at image qualities of 100 through 20 with a step size of 10.
This would yield 81 votes. The final label assigned to the sample is simply the label that got the majority votes through this process.
\\ \\
Since each of the vaccinated models is trained on a different manifold of images, the ensemble essentially models separate subspaces of the data, and the current attacks can only distort the samples in one of these subspaces. Hence, no matter which model an adversary targets, the other models should make up for the attack.
Figure \ref{fig:transferability-5} illustrates this idea.
A majority of the models are not affected significantly irrespective of the model being attacked.
Increasing JPEG compression also protects the model being attacked to some extent.
Even if the perturbation introduced is very strong, training on different compression levels help ensure that the decision boundaries learned by the vaccinated models would be dissimilar, and the verdict of the models would be highly uncorrelated.
\\ \\
We present empirical results of the accuracies obtained with the original model $M$ in Table \ref{tbl:accs-5} for comparison with our ensemble approach, where $M$ was targeted with adversarial attacks and our approach was able to recover from the attack by employing JPEG compression.
Since the ensemble involves referring to several models with varying compression levels applied to the instances being tested, a parallelized approach can also be undertaken to make the process faster.
\\ \\
Note that we choose an arbitrary combination of image qualities for our analysis, and more optimal combinations may exist.
If an adversary gains access to a classifier model and is also aware of our scheme of protecting it using this ensemble approach with vaccinated models, one can simply modify the scheme and opt for a different combination of image qualities, which would yield a completely different ensemble. Since our approach is built for faster convergence, the ensemble can be constructed quickly while still retaining the network architecture that works well for a given problem.

\section{Conclusions}
We have presented our preliminary empirical analysis of how systematic use of JPEG compression, especially in ensembles, can counter adversarial attacks and dramatically reduce their effects.
In our ongoing work, we are evaluating our approaches against more attack strategies and datasets.

\bibliographystyle{abbrv}
\bibliography{main}

\begin{thebibliography}{10}

\bibitem{bhagoji2017dimensionality}
A.~N. Bhagoji, D.~Cullina, and P.~Mittal.
\newblock Dimensionality reduction as a defense against evasion attacks on
  machine learning classifiers.
\newblock {\em arXiv preprint arXiv:1704.02654}, 2017.

\bibitem{dziugaite2016study}
G.~K. Dziugaite, Z.~Ghahramani, and D.~M. Roy.
\newblock A study of the effect of jpg compression on adversarial images.
\newblock {\em arXiv preprint arXiv:1608.00853}, 2016.

\bibitem{feinman2017detecting}
R.~Feinman, R.~R. Curtin, S.~Shintre, and A.~B. Gardner.
\newblock Detecting adversarial samples from artifacts.
\newblock {\em arXiv preprint arXiv:1703.00410}, 2017.

\bibitem{goodfellow2014explaining}
I.~J. Goodfellow, J.~Shlens, and C.~Szegedy.
\newblock Explaining and harnessing adversarial examples.
\newblock In {\em ICLR}, 2014.

\bibitem{grosse2016malware}
K.~Grosse, N.~Papernot, P.~Manoharan, M.~Backes, and P.~McDaniel.
\newblock Adversarial perturbations against deep neural networks for malware
  classification.
\newblock {\em arXiv preprint arXiv:1606.04435}, 2016.

\bibitem{gu2014towards}
S.~Gu and L.~Rigazio.
\newblock Towards deep neural network architectures robust to adversarial
  examples.
\newblock {\em arXiv preprint arXiv:1412.5068}, 2014.

\bibitem{hu2017generating}
W.~Hu and Y.~Tan.
\newblock Generating adversarial malware examples for black-box attacks based
  on gan.
\newblock {\em arXiv preprint arXiv:1702.05983}, 2017.

\bibitem{huang2017adversarial}
S.~Huang, N.~Papernot, I.~Goodfellow, Y.~Duan, and P.~Abbeel.
\newblock Adversarial attacks on neural network policies.
\newblock {\em arXiv preprint arXiv:1702.02284}, 2017.

\bibitem{IoffeS15}
S.~Ioffe and C.~Szegedy.
\newblock Batch normalization: Accelerating deep network training by reducing
  internal covariate shift.
\newblock In {\em ICML}, pages 448--456, 2015.

\bibitem{krizhevsky2009cifar10}
A.~Krizhevsky.
\newblock {Learning Multiple Layers of Features from Tiny Images}.
\newblock Master's thesis, 2009.

\bibitem{krizhevsky2012imagenet}
A.~Krizhevsky, I.~Sutskever, and G.~E. Hinton.
\newblock Imagenet classification with deep convolutional neural networks.
\newblock In {\em Advances in neural information processing systems}, pages
  1097--1105, 2012.

\bibitem{krotov2017dense}
D.~Krotov and J.~J. Hopfield.
\newblock Dense associative memory is robust to adversarial inputs.
\newblock {\em arXiv preprint arXiv:1701.00939}, 2017.

\bibitem{kurakin2016adversarial}
A.~Kurakin, I.~Goodfellow, and S.~Bengio.
\newblock Adversarial examples in the physical world.
\newblock {\em arXiv preprint arXiv:1607.02533}, 2016.

\bibitem{lin2017tactics}
Y.-C. Lin, Z.-W. Hong, Y.-H. Liao, M.-L. Shih, M.-Y. Liu, and M.~Sun.
\newblock Tactics of adversarial attack on deep reinforcement learning agents.
\newblock {\em arXiv preprint arXiv:1703.06748}, 2017.

\bibitem{luo2015foveation}
Y.~Luo, X.~Boix, G.~Roig, T.~Poggio, and Q.~Zhao.
\newblock Foveation-based mechanisms alleviate adversarial examples.
\newblock {\em arXiv preprint arXiv:1511.06292}, 2015.

\bibitem{metzen2017detecting}
J.~H. Metzen, T.~Genewein, V.~Fischer, and B.~Bischoff.
\newblock On detecting adversarial perturbations.
\newblock In {\em ICLR}, 2017.

\bibitem{Moosavi17}
S.~M. Moosavi~Dezfooli, A.~Fawzi, O.~Fawzi, and P.~Frossard.
\newblock Universal adversarial perturbations.
\newblock In {\em CVPR}, 2017.

\bibitem{Moosavi16}
S.-M. Moosavi-Dezfooli, A.~Fawzi, and P.~Frossard.
\newblock Deepfool: A simple and accurate method to fool deep neural networks.
\newblock In {\em CVPR}, 2016.

\bibitem{Papernot17blackbox}
N.~Papernot, P.~McDaniel, I.~Goodfellow, S.~Jha, Z.~B. Celik, and A.~Swami.
\newblock Practical black-box attacks against machine learning.
\newblock In {\em Proceedings of the 2017 ACM on Asia Conference on Computer
  and Communications Security}, ASIA CCS '17, pages 506--519, 2017.

\bibitem{papernot2016distillation}
N.~Papernot, P.~McDaniel, X.~Wu, S.~Jha, and A.~Swami.
\newblock Distillation as a defense to adversarial perturbations against deep
  neural networks.
\newblock In {\em IEEE Symposium on Security and Privacy}, pages 582--597,
  2016.

\bibitem{Papernot16limitation}
N.~Papernot, P.~D. McDaniel, S.~Jha, M.~Fredrikson, Z.~B. Celik, and A.~Swami.
\newblock The limitations of deep learning in adversarial settings.
\newblock In {\em {IEEE} European Symposium on Security and Privacy, EuroS{\&}P
  2016, Saarbr{\"{u}}cken, Germany, March 21-24, 2016}, pages 372--387, 2016.

\bibitem{PapernotMSH16}
N.~Papernot, P.~D. McDaniel, A.~Swami, and R.~E. Harang.
\newblock Crafting adversarial input sequences for recurrent neural networks.
\newblock In {\em 2016 {IEEE} Military Communications Conference, {MILCOM}},
  pages 49--54, 2016.

\bibitem{Stallkamp-IJCNN-2011}
J.~Stallkamp, M.~Schlipsing, J.~Salmen, and C.~Igel.
\newblock The {G}erman {T}raffic {S}ign {R}ecognition {B}enchmark: A
  multi-class classification competition.
\newblock In {\em IEEE International Joint Conference on Neural Networks},
  pages 1453--1460, 2011.

\bibitem{Szegedy14}
C.~Szegedy, G.~Inc, W.~Zaremba, I.~Sutskever, G.~Inc, J.~Bruna, D.~Erhan,
  G.~Inc, I.~Goodfellow, and R.~Fergus.
\newblock Intriguing properties of neural networks.
\newblock In {\em ICLR}, 2014.

\end{thebibliography}

\end{document}